\documentclass{article}

\usepackage{graphicx,amsmath,bm}

\title{Analysis of dropout learning regarded as ensemble learning}
\author{Kazuyuki Hara$^{1}$ \and Daisuke Saitoh$^{2}$ \and Hayaru Shouno$^{3}$ %
}
\date{College of Industrial Technology, Nihon University,\\1-2-1 
Izumi-cho, Narashino-shi, Chiba, 275-8575 Japan. \\
Graduate School of Industrial Technology, Nihon University\\ 
Graduate School of  Informatics and Engineering, \\The University of Electro-Communications\\
1-5-1 Chofugaoka, Chofu-shi, Tokyo, 182-8585 Japan.
}
\begin{document}
\maketitle
\begin{abstract}
Deep learning is the state-of-the-art in fields such as visual object recognition and speech recognition. This learning uses a large number of layers, huge number of units, and connections. Therefore, overfitting is a serious problem. To avoid this problem, dropout learning is proposed. Dropout learning neglects some inputs and hidden units in the learning process with a probability, $p$, and then, the neglected inputs and hidden units are combined with the learned network to express the final output. We find that the process of combining the neglected hidden units with the learned network can be regarded as ensemble learning, so we analyze dropout learning from this point of view. 
\end{abstract}

{\bf keywords: }Dropout learning, overfitting, regularization, ensemble learning, soft-committee machine, teacher-student formulation \\

\section{Introduction}
Deep learning \cite{Hinton2006,LeCun2015} is attracting much attention in the field of visual object recognition, speech recognition, object detection, and many other domains. It provides automatic feature extraction and has the ability to achieve outstanding performance \cite{Hinton2012a,Deng2013}. 

Deep learning uses a very deep layered network and a huge number of data, so overfitting is a serious problem. To avoid overfitting, regularization is used. Hinton et al. proposed a regularization method called ``dropout learning" \cite{Hinton2012} for this purpose. Dropout learning follows two processes. At learning time, some hidden units are neglected with a probability $p$, and this process reduces the network size. At test time, learned hidden units and those not learned are summed up and multiplied by $p$ to calculate the network output. We find that summing up the learned and not learned units multiplied by $p$ can be regarded as ensemble learning. 

In this paper, we analyze dropout learning regarded as ensemble learning \cite{hara2005}. On-line learning \cite{Biehl1995,Saad1995} is used to learn a network. We analyze dropout learning regarded as ensemble learning, except for using different sets of of hidden units in dropout learning. We also analyze dropout learning regarded as an L2 normalizer \cite{Wager2014}. 

\section{Model}
\label{model}
In this paper, we use a teacher-student formulation and assume the existence of a teacher network (teacher) that produces the desired output for the student network (student). By introducing the teacher, we can directly measure the similarity of the student weight vector to that of the teacher. First, we formulate a teacher and a student, and then introduce the gradient descent algorithm. 

\begin{figure}[h]
\begin{center}
\includegraphics[width=7cm]{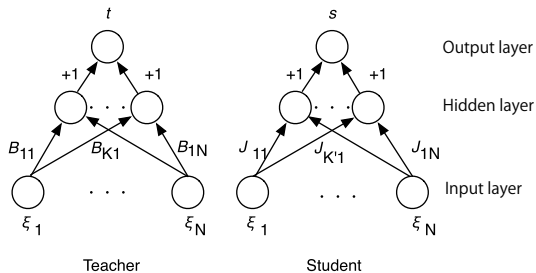}
\end{center}
\vspace{-0.5cm}
\caption{\label{network}Network structures of teacher and student}
\end{figure}

The teacher and student are a soft committee machine with $N$ input units, hidden units, and an output, as shown in Fig. \ref{network}. The teacher consists of $K$ hidden units, and the student consists of $K'$ hidden units. Each hidden unit is a perceptron. The $k$th hidden weight vector of the teacher is $\bm{B}_k=(B_{k1}, \ldots ,B_{kN})$, and the $k'$th hidden weight vector of student is $\bm{J}_{k'}^{(m)} = (J_{k'1}^{(m)}, \ldots , J_{k'N}^{(m)})$, where $m$ denotes learning iterations. In the soft committee machine, all hidden-to-output weights are fixed to be $+1$ \cite{Saad1995}. This network calculates the majority vote of hidden outputs. 

\vspace{-0.5cm}

We assume that both the teacher and the student receive $N$-dimensional input $\bm{\xi}^{(m)}=(\xi_1^{m}, \ldots, \xi_N^{(m)})$, that the teacher outputs $t^{(m)}=\sum_{k=1}^{K}t_k^{(m)}=\sum_{k=1}^{K}g(d_k^{(m)})$, and that the student outputs $s^{(m)}=\sum_{k'=1}^{K'} s_{k'}^{(m)}=\sum_{k'=1}^{K'} g(y_{k'}^{(m)})$. Here, $g(\cdot)$ is the output function of a hidden unit, $d_k^{(m)}$ is the inner potential of the $k$th hidden unit of the teacher calculated using $d_k^{(m)}=\sum_{i=1}^N B_{ki} \xi_i^{(m)}$, and $y_{k'}^{(m)}$ is the inner potential of the $k'$th hidden unit of the student calculated using $y_{k'}^{(m)}=\sum_{i=1}^N J_{k'i}^{(m)} \xi_{i}^{(m)}$. 

We assume that the $i$th elements $\xi_i^{(m)}$ of the independently drawn input $\bm{\xi}^{(m)}$ are uncorrelated random variables with zero mean and unit variance; that is, that the $i$th element of the input is drawn from a probability distribution $\mbox{P}(\xi_i)$. The thermodynamic limit of $N \rightarrow \infty$ is also assumed. The statistics of the inputs in the thermodynamic limit are $\left<\xi_{i}^{(m)}\right> =0$, $\left<(\xi_{i}^{(m)})^2\right>\equiv \sigma_{\xi}^2=1$, and $\left<\|\bm{\xi}^{(m)}\|\right>=\sqrt{N}$, where $\left< \cdots \right>$ denotes the average and $\| \cdot \|$ denotes the norm of a vector. Each element $B_{ki}, \ k=1 \sim K$ is drawn from a probability distribution with zero mean and $1/N$ variance. With the assumption of the thermodynamic limit, the statistics of the teacher weight vector are $\left<B_{ki}\right>=0, \left< (B_{ki})^2\right> \equiv \sigma_{B}^2=1/N$, and $\left<\| \bm{B_k}\|\right>=1$. This means that any combination of $\bm{B}_{l} \cdot \bm{B}_{l'}=0$. The distribution of inner potential $d_{k}^{(m)}$ follows a Gaussian distribution with zero mean and unit variance in the thermodynamic limit. 

For the sake of analysis, we assume that each element of $J_{k'i}^{(0)}$, which is the initial value of the student vector $\bm{J}_{k'}^{(0)}$, is drawn from a probability distribution with zero mean and $1/N$ variance. The statistics of the $k'$th hidden weight vector of the student are $\left< J_{k'i}^{(0)}\right>=0, \left<(J_{k'i}^{(0)})^2\right>\equiv \sigma_{J}^2=1/N$, and $\left<\|\bm{J}_{k'}^{(0)}\|\right>=1$ in the thermodynamic limit. This means that any combination of $\bm{J}_{l}^{(0)} \cdot \bm{J}_{l'}^{(0)}=0$. The output function of the hidden units of the student $g(\cdot)$ is the same as that of the teacher. The statistics of the student weight vector at the $m$th iteration are $\left< J_{k'i}^{(m)}\right>=0$, $\left<(J_{k'i}^{(m)})^2\right> =(Q_{k'k'}^{(m)})^2/N$, and $\left<\|\bm{J}_{k'}^{(m)}\|\right>=Q_{k'k'}^{(m)}$. Here, $(Q_{k'k'}^{(m)})^2=\bm{J}_{k'}^{(m)}\cdot \bm{J}_{k'}^{(m)}$. The distribution of the inner potential $y_{k'}^{(m)}$ follows a Gaussian distribution with zero mean and $(Q_{k'k'}^{(m)})^2$ variance in the thermodynamic limit. 

Next, we introduce the stochastic gradient descent (SGD) algorithm for the soft committee machine. 
The generalization error is defined as the squared error $\varepsilon$ averaged over possible inputs: 

\begin{equation}
\varepsilon_g^{(m)}=\left<\varepsilon^{(m)} \right>=\frac{1}{2}\left< (t^{(m)}-s^{(m)})^2 \right>=\frac{1}{2}\left< \left(\sum_{k=1}^Kg(d_k^{(m)})-\sum_{k'=1}^{K'}g(y_{k'}^{(m)})\right)^2 \right> ,\label{eg}
\end{equation}

At each learning step $m$, a new uncorrelated input, $\bm{\xi}^{(m)}$, is presented, and the current hidden weight vector of the student $\bm{J}_{k'}^{(m)}$ is updated using

\begin{equation}
\bm{J}_{k'}^{(m+1)}=\bm{J}_{k'}^{(m)}+\frac{\eta}{N}\left(\sum_{l=1}^K g(d_l^{(m)})-\sum_{l'=1}^{K'}g(y_{l'}^{(m)})\right)g^{\prime}(y_{k'}^{(m)})\bm{\xi}^{(m)}, \label{le}
\end{equation}

\noindent 
where $\eta$ is the learning step size and $g^{\prime}(x)$ is the derivative of the output function of the hidden unit $g(x)$. 

On-line learning uses a new input at once, therefore, overfitting does not occur. To evaluate the dropout learning in on-line learning, pre-selected whole inputs frequently use in a on-line manner. From our experiences, when the input dimension is $N$, then overfitting occurs for pre-selected whole $10\times N$ inputs.
  
\section{Dropout learning and ensemble learning}
In this section, we compare dropout learning and ensemble learning regarded as a way of calculating network output. 
\subsection{Ensemble learning}
Eensemble learning is performed by using many learners (referred to as students) to achieve better performance \cite{hara2005}. In ensemble learning, each student learns the teacher independently, and each output is averaged to calculate the ensemble output $s_{en}$. 

\begin{equation}
s_{en}=\sum_{k'_{en}=1}^{K_{en}} C_{k'_{en}}s_{k'_{en}}=\sum_{k'_{en}=1}^{K_{en}} C_{k'_{en}}\sum_{k'=1}^{K'} g(y_{k'})
\label{ensemble_output}
\end{equation}

\noindent
Here, $C_{k'_{en}}$ is a weight for averaging. $K_{en}$ is the number of students. 

\begin{figure}[h]
\begin{minipage}{0.6\hsize}
\includegraphics[width=6cm]{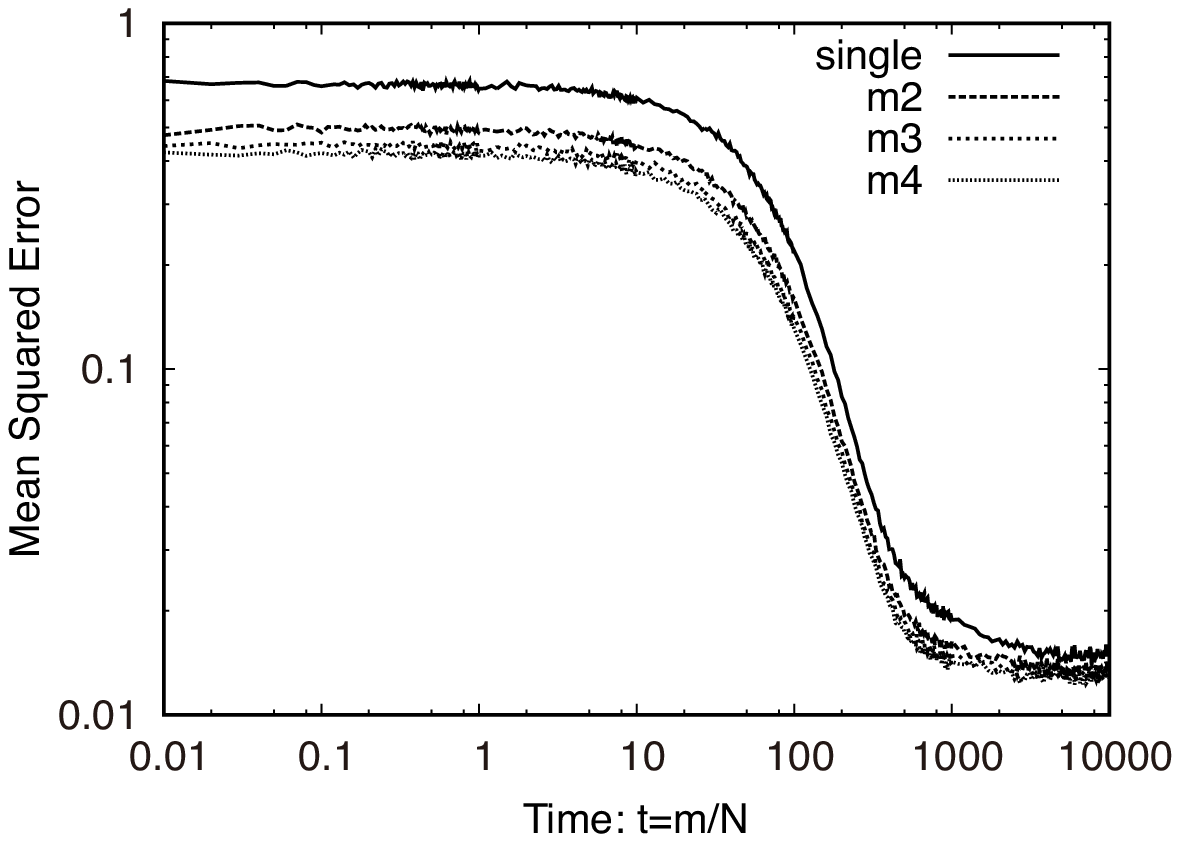}
\caption{\label{ensemble}Effect of ensemble learning }
\end{minipage}
\hspace{-0.7cm}
\begin{minipage}{0.4\hsize}
\begin{center}
\includegraphics[width=5.5cm]{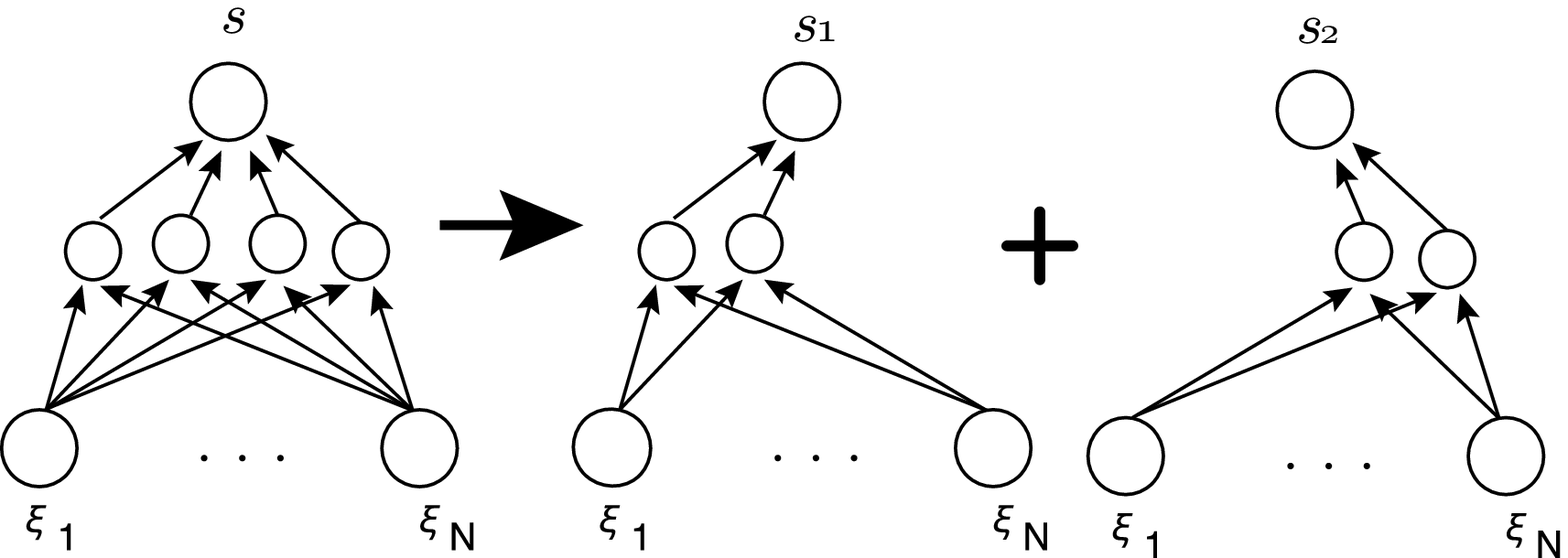}
\end{center}
\caption{\label{ensemble2}Network divided into two networks to apply ensemble learning }
\end{minipage}
\end{figure}

Figure \ref{ensemble} shows computer simulation results. The teacher and student include two hidden units. 
The output function $g(x)$ is the error function $\mbox{erf}(x/\sqrt{2})=\int_{-x}^{x} dt \exp(-t^2/s)/\sqrt{2\pi}$.  
In the figure, the horizontal axis is time $t=m/N$. Here, $m$ is the iteration number, and $N$ is the dimension of input units. Input dimension is $N=10000$, and $10\times N$ inputs are frequently used.  The vertical axis is the mean squared error (MSE) for $N$ input data. Each elements $\xi_i^{(m)}$ of the independently drawn input $\bm{\xi}^{(m)}$ are uncorrelated random variables with zero mean and unit variance. Target for $\bm{\xi}^{(m)}$ is the teacher output. 
The teacher and the initial student weight vectors are set as described in Sec. \ref{model}. In the figure, ``Single" is the result of using a single student. ``m2" is the result of using an ensemble of two students, ``m3" is that of an ensemble of three students, and ``m4" is that of  ensemble of four students.  As shown, the ensemble of four students outperformed the other two cases. 


\vspace{-0.5cm}

Next, we modify the ensemble learning. We divide the student  (with $K'$ hidden units) into $K_{en}$ networks (See Fig. \ref{ensemble2}. Here, $K'=4$ and $K_{en}=2$). These divided networks learn the teacher independently, and then we calculate the ensemble output $s_{en}$ by averaging the outputs $s_{k'_{en}l'}$ as: 

\begin{equation}
s_{en}=\frac{1}{K_{en}}\sum_{k'_{en}=1}^{K_{en}} s_{k'_{en}}=\frac{1}{K_{en}}\sum_{k'_{en}=1}^{K_{en}} \sum_{l'=1}^{M/K_{en}} g(y_{k'_{en}l'}).
\label{ensemble_learning}
\end{equation}

\noindent
Here, $s_{k'_{en}}$ is the output of a divided network with $M/K_{en}$ hidden units, and $g(y_{k'_{en}l'})$ is the $l'$th hidden output in the $k'_{en}$th divided network. Eq. (\ref{ensemble_learning}) corresponds to Eq. ({\ref{ensemble_output}) when $C_{k'_{en}}=\frac{1}{K_{en}}$ and $K'=\frac{M}{K_{en}}$.

\subsection{Dropout learning}
\label{dropout}
In this subsection, we introduce dropout learning \cite{Hinton2012}. Dropout learning is used in deep learning to prevent overfitting. A small number of data compared with the size of a network may cause overfitting \cite{Bishop}. In the state of overfitting, the learning error (the error for learning data) and the test error (the error by cross-validation) become different. Figure \ref{example} shows the result of the SGD and that of dropout learning. The soft committee machine was used for both the teacher and student. $\mbox{erf}(x/\sqrt{2})$ was used as the output function $g(x)$. Input dimension is $N=1000$, and the teacher had two hidden units, and the student had 100 hidden units. The input and its target are generated as those of Fig.\ref{ensemble}. The learning step size $\eta$ was set to $0.01$, and $1000$ pieces of inputs were used iteratively for learning. In Fig. \ref{example}(a) shows the learning curve of the SGD. In this setting, overfitting occurred. Figure \ref{example}(b) shows the learning curve of the SGD with dropout learning. The learning error was small compared with the test error; however, the difference between the learning error and the test error was not as significant as that of the SGD. Therefore, these results shows that dropout learning prevent overfitting.

\begin{figure}[h]
\includegraphics[width=6cm]{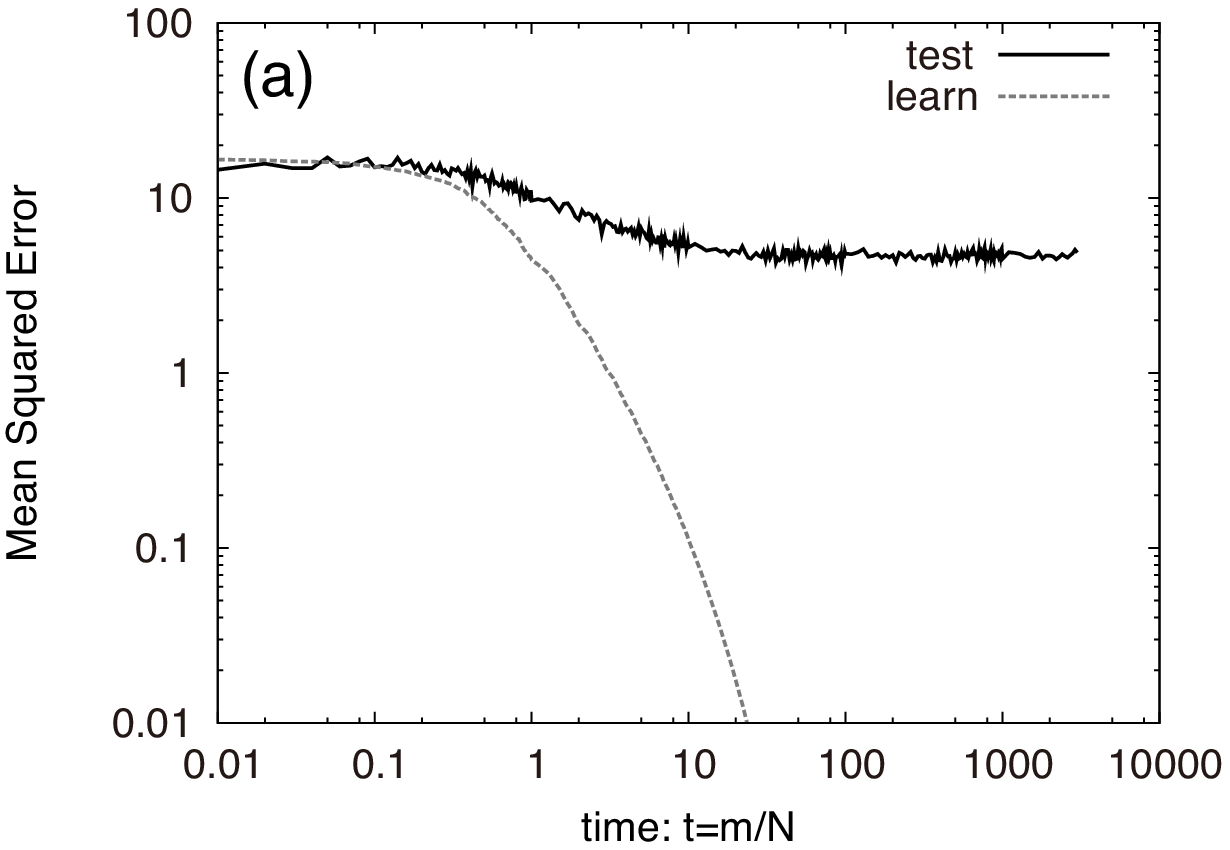}
\includegraphics[width=6cm]{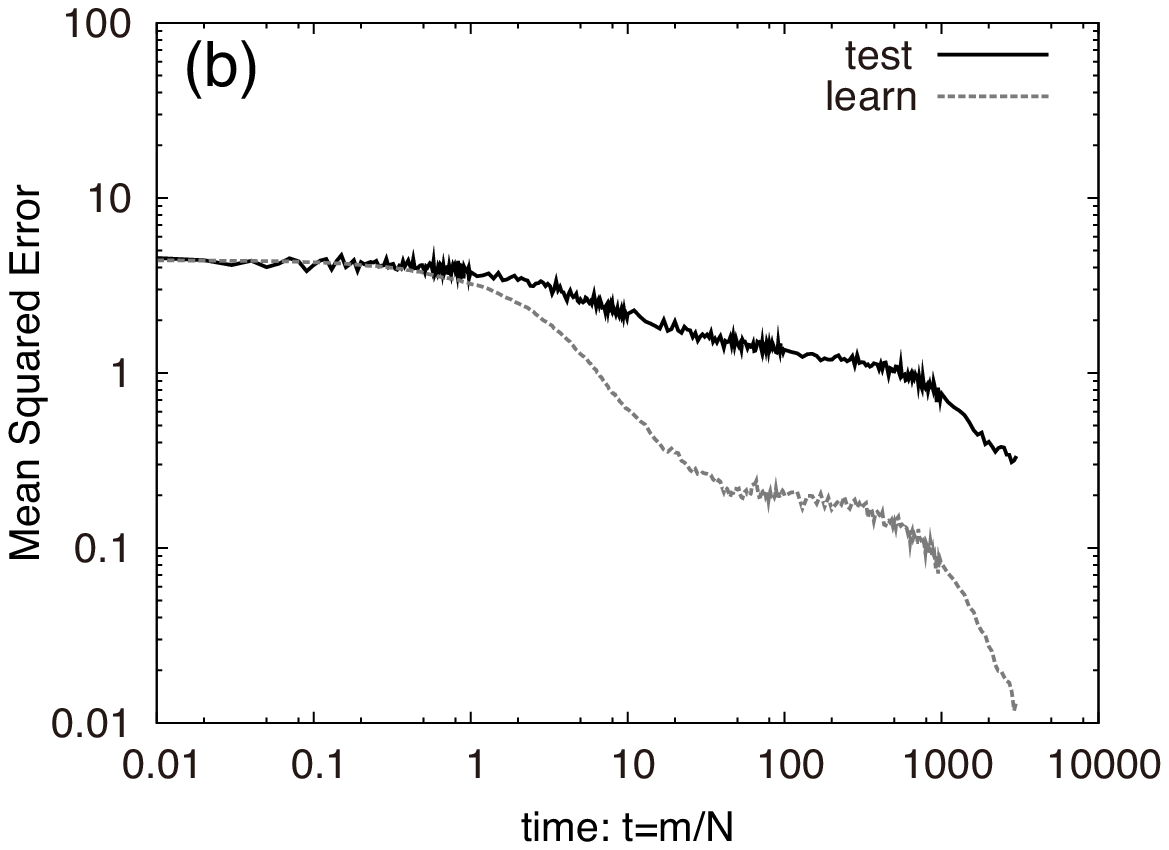}
\vspace{-0.2cm}
\caption{\label{example}Effect of dropout. (a) is learning curve of SGD, and (b) is that of dropout learning.}
\end{figure}

\vspace{-0.5cm}

The learning equation of dropout learning for the soft committee machine can be written as the next equation.

\begin{equation}
\bm{J}_{k'}^{(m+1)}=\bm{J}_{k'}^{(m)}+\frac{\eta}{N}\left(\sum_{l=1}^K g(d_l^{(m)})-\sum_{l'\notin D^{(m)}}^{(1-p)K'}g(y_{l'}^{(m)})\right)g^{\prime}(y_{k'}^{(m)})\bm{\xi}^{(m)}, \label{le_drop}
\end{equation}

\noindent
Here, $D^{(m)}$ shows a set of hidden units that is randomly selected with respect to the probability $p$ from all the hidden units at the $m$th iteration. The hidden units in $D^{(m)}$ are not subject to learning. After the learning, the student's output $s^{(m)}$ is calculated by the sum of learned hidden outputs and those not learned multiplied by $p$.

\begin{equation}
s^{(m)}=p*\left\{\sum_{l'\notin {D}^{(m)}}^{(1-p)K'} g(y_{l'}^{(m)})+\sum_{l' \in D^{(m)}}^{pK'} g(y_{l'}^{(m-1)})\right\}
\label{dropout_learning}
\end{equation}

\noindent
This equation is regarded as the ensemble of a learned network (the first term) and that of a not learned network (the second term) when the probability is $p=0.5$. Equation \ref{dropout_learning} is correspond to Eq. (\ref{ensemble_learning}) when $p=1/K_{en}$ and $K_{en}=2$. However, a set of hidden units in $D^{(m)}$ is selected at random in every iteration. So, dropout learning is regarded as  ensemble learning performed by using a different set of hidden units in every iteration. Instead, the original ensemble learning is the average of the fixed set of hidden units throughout the learning. This difference may cause the difference in performances between dropout learning and ensemble learning. 

\section{Results}
\subsection{Comparison between dropout learning and ensemble learning}
In this section, the error function $\mbox{erf}(x/\sqrt{2})$ is used as the output function $g(x)$. We compared dropout learning and ensemble learning. We used two soft committee machines with 50 hidden units for ensemble learning. For dropout learning, we used one soft committee machine with 100 hidden units. We set $p=0.5$; then, dropout learning selected 50 hidden units in $D^{(m)}$ with 50 unselected hidden units remaining. Therefore, dropout learning and ensemble learning had the same architectures. Input dimension is $N=1000$, and the learning step size was set to $\eta=0.01$. The input and its target are generated as those of Fig.\ref{ensemble}. $N$ inputs were used iteratively for learning. Figure \ref{drop-and-ensemble} shows the results. The horizontal axes is time $t=m/N$, and the vertical axis is the MSE calculated for $N$ input data. In Fig. \ref{drop-and-ensemble}(a), ``single" shows the soft-committee machines with 50 hidden units. ``ensemble" shows the results given by ensemble learning. Test errors are used in these figures. In Fig. \ref{drop-and-ensemble}(b), ``test" shows the MSE given by the test data. ``learn" shows the MSE given by the learning data. Results are obtained by average of 10 trials. As shown in Fig. \ref{drop-and-ensemble}(a), the ensemble learning achieved an MSE smaller than that of the single network. However, dropout learning achieved an MSE smaller than that of ensemble learning. Therefore, ensemble learning using a different set of hidden units in every iteration (this is the dropout) performs better than when using the same set of hidden units throughout the learning. Note that even with dropout learning using more hidden units than ensemble learning, overfitting did not occur. Therefore, in the next subsection, we will compare dropout learning with the SGD with L2 regularization.

\begin{figure}[h]
\includegraphics[width=6cm]{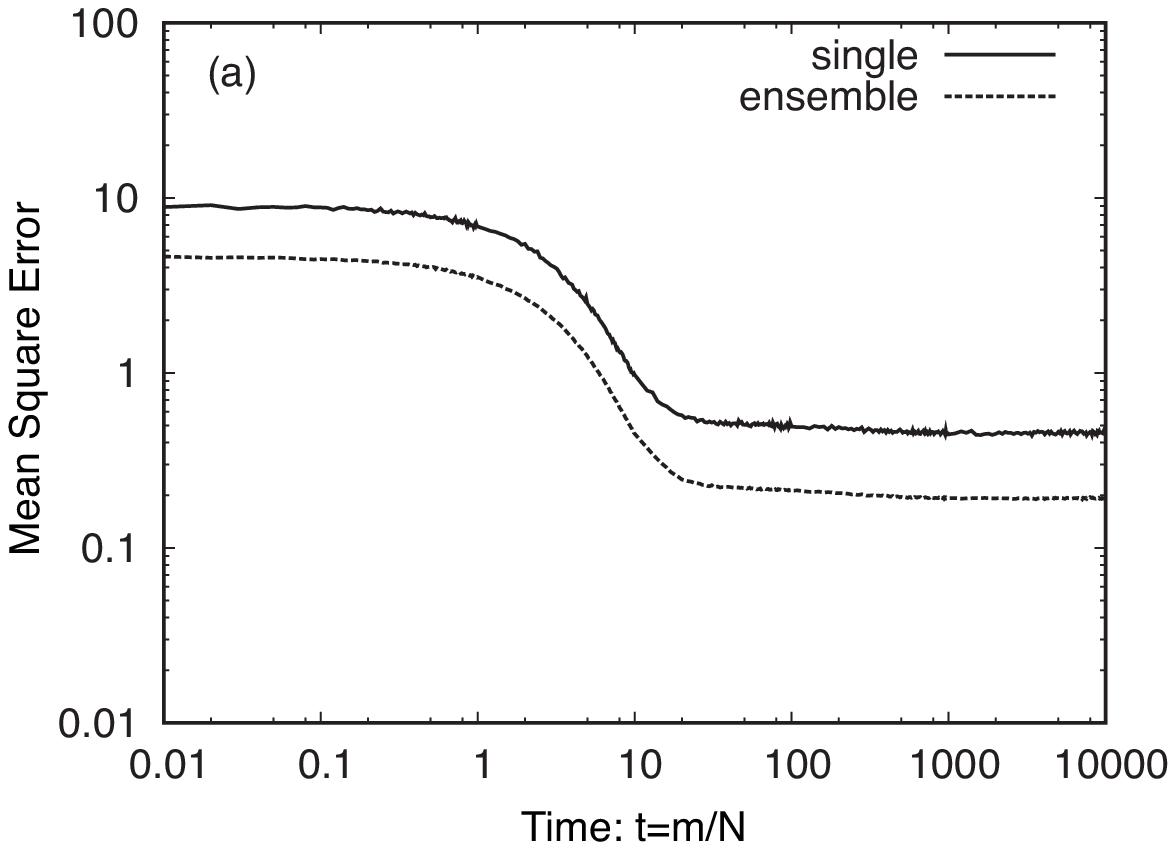}
\includegraphics[width=6.2cm,height=4.5cm]{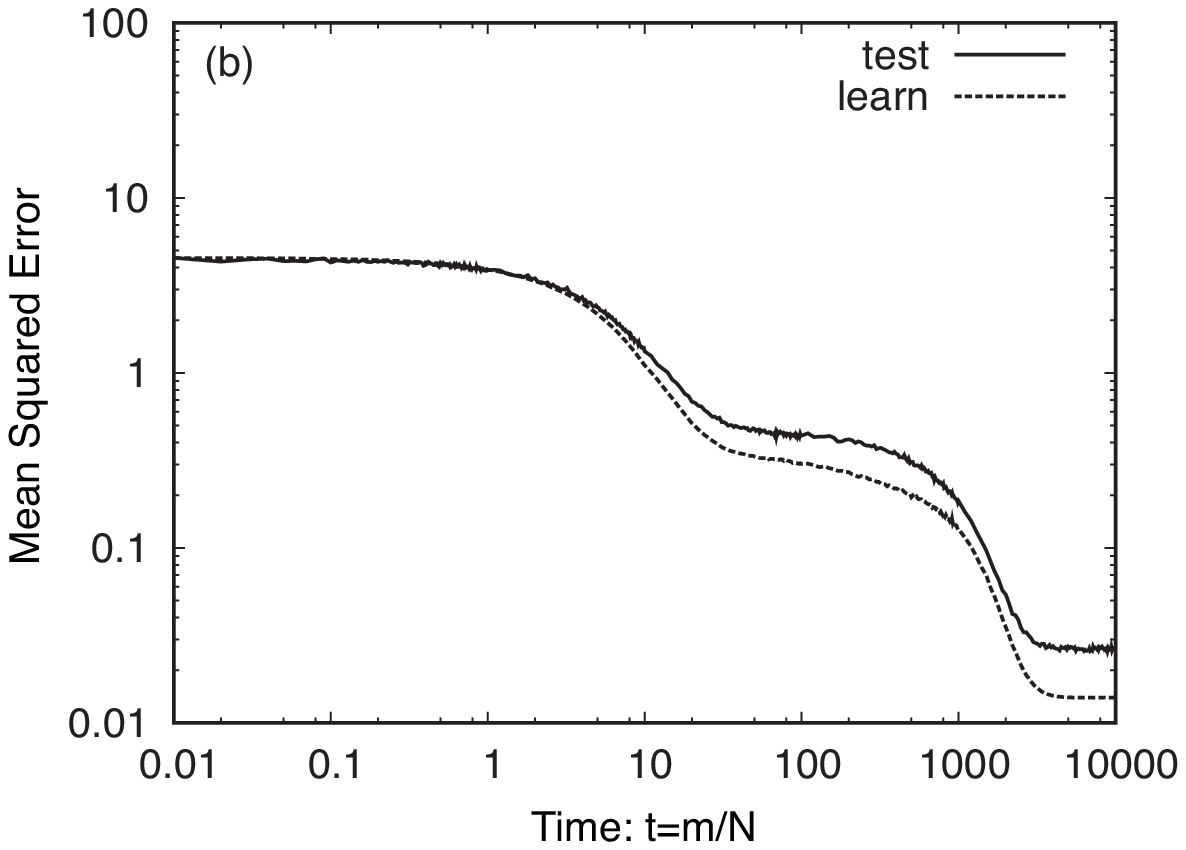}
\vspace{-0.2cm}
\caption{\label{drop-and-ensemble}Results of comparison between dropout learning and ensemble learning. (a) is ensemble learning of two networks, and (b) is dropout learning with respect to $p=0.5$. }
\end{figure}

\vspace{-0.2cm}

\subsection{Comparison between dropout learning and SGD with L2 regularization}
The next learning equation shows the SGD with L2 regularization.

\begin{equation}
\bm{J}_{k'}^{(m+1)}=\bm{J}_{k'}^{(m)}+\frac{\eta}{N}\left(\sum_{l=1}^K g(d_l^{(m)})-\sum_{l'=1}^{K'}g(y_{l'}^{(m)})\right)g^{\prime}(y_{k'}^{(m)})\bm{\xi}^{(m)}-\alpha \|\bm{J}_{k'}^{(m)}\|^2. \label{le3}
\end{equation}

\noindent
Here, $\alpha$ is a coefficient of the L2 penalty. 

In Fig. \ref{L2}, we show the learning results of the SGD with L2 regularization. 
Results are obtained by average of 10 trials. The conditions were the same as those of Fig. \ref{drop-and-ensemble}. 

\begin{figure}[h]
\begin{center}
\includegraphics[width=6cm]{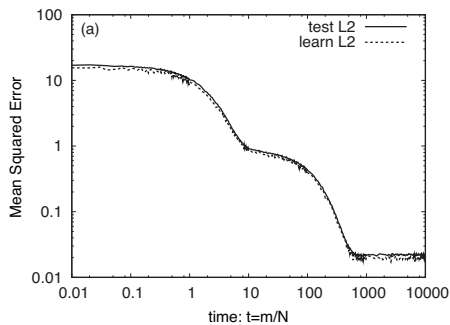}
\end{center}
\vspace{-0.2cm}
\caption{\label{L2}Learning curve of SGD with L2 normalization}
\end{figure}


\noindent
From comparison between Fig. \ref{L2} and Fig. \ref{drop-and-ensemble}(b), the residual error of dropout learning was almost the same as that of the SGD with L2 regularization. Therefore, the regularization effort of dropout learning is the same as the L2 regularization. Note that for the SGD with L2 regularization, we must choose $\alpha$ in trials; however, dropout learning has no tuning parameter. 

\section{Conclusion}
In this paper, we analyzed dropout learning regarded as ensemble learning. In ensemble learning, we divide the network into several sub-networks, and then we learn each sub-network independently. After the learning, the ensemble output is calculated by using the average of the sub-network outputs. We showed that dropout learning can be regarded as ensemble learning except for using a different set of hidden units in every learning iteration. Using a different set of hidden unit outperforms ensemble learning. 
We also showed that dropout learning achieves the same performance as the L2 regularizer. Our future work is the theoretical analysis of dropout learning with ReLU activation function.

\subsection*{Acknowledgments}
The authors thank Dr.  Masato Okada and Dr. Hideitsu Hino for insightful discussions.

\end{document}